\documentclass[sigconf]{acmart}
\AtBeginDocument{%
  }

\copyrightyear{2025}
\acmYear{2025}
\setcopyright{acmlicensed}\acmConference[KDD '25]{Proceedings of the 31st ACM SIGKDD Conference on Knowledge Discovery and Data Mining V.2}{August 3--7, 2025}{Toronto, ON, Canada}
\acmBooktitle{Proceedings of the 31st ACM SIGKDD Conference on Knowledge Discovery and Data Mining V.2 (KDD '25), August 3--7, 2025, Toronto, ON, Canada}
\acmDOI{10.1145/3711896.3737051}
\acmISBN{979-8-4007-1454-2/2025/08}

\usepackage{amsmath}
\usepackage{amsfonts}
\usepackage{amsthm}
\usepackage{booktabs}
\usepackage{xspace}
\usepackage{subfigure}
\usepackage{pifont}
\usepackage{mathtools}
\usepackage{environ}
\usepackage{multirow}
\usepackage{pdfpages}
\usepackage{algorithm}
\usepackage{algorithmic}
\newcommand{\model}{MockLLM\xspace}

\newcommand{\aka}{\emph{a.k.a.,}\xspace}
\newcommand{\eg}{\emph{e.g.,}\xspace}

\begin{document}

\title{MockLLM: A Multi-Agent Behavior Collaboration Framework for Online Job Seeking and Recruiting}


\author{Hongda Sun}\authornote{Equal contribution.}
\affiliation{
  \institution{Gaoling School of Artificial Intelligence, Renmin University of China}
  \city{Beijing}
  \country{China}
}
\email{sunhongda98@ruc.edu.cn}

\author{Hongzhan Lin}\authornotemark[1]
\affiliation{
  \institution{Engineering Research Center of Next-Generation Intelligent Search and Recommendation, Ministry of Education}
  \city{Beijing}
  \country{China}
}
\email{linhongzhan@ruc.edu.cn}

\author{Haiyu Yan}
\affiliation{
  \institution{University of Chinese Academy of Sciences}
  \city{Beijing}
  \country{China}
}
\email{yanhaiyu24@mails.ucas.ac.cn}

\author{Yang Song}
\affiliation{
  \institution{Nanbeige Lab, BOSS Zhipin}
  \city{Beijing}
  \country{China}
}
\email{songyang@kanzhun.com}

\author{Xin Gao}
\affiliation{
  \institution{King Abdullah University of Science and Technology}
  \city{Thuwal}
  \country{Saudi Arabia}
}
\email{xin.gao@kaust.edu.sa}

\author{Rui Yan}\authornote{Corresponding author: Rui Yan (\url{ruiyan@ruc.edu.cn}).}
\affiliation{
  \institution{Gaoling School of Artificial Intelligence, Renmin University of China}
  \institution{School of Artificial Intelligence, Wuhan University}
  \city{Beijing and Wuhan}
  \country{China}
}
\email{ruiyan@ruc.edu.cn}

\renewcommand{\shortauthors}{Hongda Sun et al.}

\begin{abstract}
Online recruitment platforms have reshaped job-seeking and recruiting processes, driving increased demand for applications that enhance person-job matching. Traditional methods generally rely on analyzing textual data from resumes and job descriptions, limiting the dynamic, interactive aspects crucial to effective recruitment. Recent advances in Large Language Models (LLMs) have revealed remarkable potential in simulating adaptive, role-based dialogues, making them well-suited for recruitment scenarios. In this paper, we propose \textbf{MockLLM}, a novel framework to generate and evaluate mock interview interactions. The system consists of two key components: mock interview generation and two-sided evaluation in handshake protocol. By simulating both interviewer and candidate roles, MockLLM enables consistent and collaborative interactions for real-time and two-sided matching. To further improve the matching quality, MockLLM further incorporates reflection memory generation and dynamic strategy modification, refining behaviors based on previous experience. We evaluate MockLLM on real-world data Boss Zhipin, a major Chinese recruitment platform. The experimental results indicate that MockLLM outperforms existing methods in matching accuracy, scalability, and adaptability across job domains, highlighting its potential to advance candidate assessment and online recruitment.

\end{abstract}




\maketitle

\section{Introduction}

With the emergence of online services, the job market has undergone continuous revolution, offering job seekers and recruiters greater opportunities to connect across geographical boundaries.
The services have facilitated the acquisition of real-world interview data, further increasing the need to develop higher-quality applications to enhance person-job matching.

Since candidate and recruiter information is often represented through textual resumes and job descriptions, most existing studies focus on modeling the latent semantics of these documents and developing a matching function to measure their suitability~\cite{pjfnn,apjfnn}.
Given the remarkable capabilities of Large Language Models (LLMs)~\cite{instructgpt,qin2023chatgpt,gpt4}, particularly in various role-playing scenarios~\cite{sun2024harnessing,sun2024determlr,liu2025mobilesteward}, it is worth exploring the potential of LLMs to simulate critical roles to form conversational mock interviews. 
These mock interview questions and answers can serve as augmented data, which provides additional evidence for evaluating recruiters and job seekers, thereby enhancing better matching outcomes for both sides.

However, previous role-playing frameworks typically assign each LLM to specialize in a single task, optimizing its performance in that specific area~\cite{chen2023autoagents,chen2023agentverse}.
While in online recruitment, the same interviewer must conduct both the interview and the post-interview evaluation to ensure consistency, and the same applies to the candidate role. Therefore, it is natural for both roles to behave in multiple functions with collaborations.
Moreover, building high-quality mock interviews is critical yet challenging, which requires high professionalism from interviewers to ask appropriate questions and make overall assessments. Similarly, it is also challenging for job seekers to provide decent answers and job hunting decisions.

\begin{figure}
    \centering
    \includegraphics[width=\linewidth]{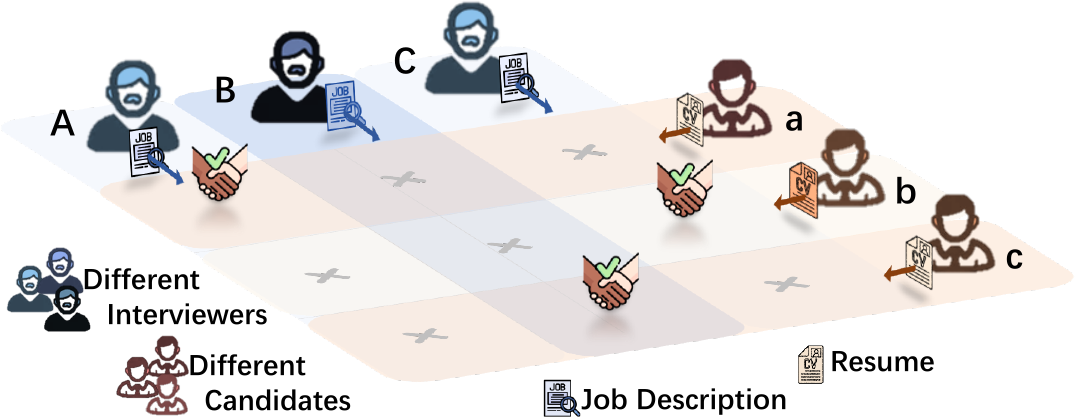}
    \caption{An example platform for two-sided handshake evaluation protocol between interviewers and candidates.}
    \label{fig:intro}
\end{figure}

To address these challenges, we propose \model, a novel framework that facilitates multi-agent behavior collaboration for person-job matching.
Unlike traditional role-playing frameworks where each LLM agent is restricted to a single function, \model allows for multiple behaviors for LLM-simulated interviewers and job seekers, enhancing their capabilities to handle various sub-tasks. 
The behaviors primarily undertaken by interviewers in our framework include: (1) \textit{raising interview questions} during the interview stage: eliciting key information from the candidate's resume and asking valuable questions via a multi-turn interview process; (2) \textit{evaluating interview performance} during the review stage: 
assessing the appropriateness of candidates for the position based on their interview performance. On the other side, candidates engage with the following behaviors: (1) \textit{answering interview questions}: providing qualified responses to the interview questions; (2) \textit{evaluating job positions}: assessing their interest and confidence in accepting a job offer.
As illustrated in Figure~\ref{fig:intro}, both interviewers and candidates need to evaluate multiple options on the platform.
To ensure more effective two-sided matching, we introduce a \textit{handshake protocol} in the review stage, integrating evaluation results from both sides to make final matching decisions.
By following the handshake protocol, both parties can achieve a mutual match, improving the overall quality of person-job fitting. 

To continuously improve the quality of mock interviews, we propose \textit{reflection memory generation} for both interviewers and candidates to memorize valuable interview experiences and refine interviewing behaviors.
Once a two-sided match is achieved in the handshake protocol, the interview questions and answers are regarded as positive outcomes for future reference. Interviewers can use these matched cases to better engage with new candidates with similar backgrounds, while candidates use them to respond to similar job-related questions. The reflection memory contributes to enhancing the strategies used in future interviews.

We conduct comprehensive experiments based on real-world talent recruitment data from Boss Zhipin, a leading online hiring platform in China.
The results demonstrate that MockLLM consistently outperforms traditional models in person-job matching accuracy, scalability, and cross-domain adaptability, underscoring its robustness and utility. Additionally, simulated online testing highlights MockLLM’s potential to drive future advancements in real-world online recruitment deployment.

To sum up, our contributions can be summarized as follows:

$\bullet$ We propose \model, a novel framework that develops role-playing LLMs to conduct mock interviews, providing augmented evidence for better evaluation of person-job fitting.

$\bullet$ We design a multi-agent behavior collaboration paradigm in \model, enabling a single LLM to master multiple behaviors in mock interview generation and two-sided evaluation.

$\bullet$ We introduce dynamic strategy modification techniques to refine the behaviors of interviewers and candidates, continuously improving mock interview quality to enhance person-job matching.

\section{Related Work}
\subsection{Person-Job Matching}
Recruitment-oriented talent science research has consistently held a central position in human resource management, contributing to the prosperous growth of both enterprises and talents~\cite{harris2017finding,rendle2012bpr}.
At the core of this research domain lies the task of person-job matching, which aims to align job seekers with appropriate job opportunities based on their qualifications, preferences, and potential fit~\cite{zhu2018person}. Over the years, this area has seen continuous advancement, with a growing emphasis on applying machine learning and recommendation techniques to improve matching outcomes~\cite{li2024does}.
Malinowski et al. present a recommendation model in the recruitment scenario through a personalized search for candidates~\cite{malinowski2006matching}. 
Modern approaches increasingly frame person-job matching as a recommendation problem.
NCF is proposed as a collaborative filtering based method that utilizes multilayer perception to simulate the interaction between jobs and candidates~\cite{ncf}.
PJFNN employs two convolutional neural networks to encode jobs and resumes separately, and then compute cosine similarity based matching scores~\cite{pjfnn}. APJFNN applies recurrent neural networks as document encoders and integrates attention mechanisms to model work abilities and skills~\cite{apjfnn}.
IPJF leverages multiple labels to indicate the propensity of candidates and jobs to reach a match~\cite{ipjf}.

In addition to job descriptions and resume texts, some research has found that it could bring benefits by incorporating interview history into person-job matching. 
Yan et al. combine the interview history of job seekers and recruiters into the memory network to help improve person-job matching~\cite{yan2019interview}.
JLMIA leverages an unsupervised learning framework to integrate interview question generation and person-job fit using latent topic discovery~\cite{jlmia}.
More recently, EZInterviewer improves job interview quality using a knowledge-grounded dialogue model to connect the multi-turn interview session with candidates' resumes~\cite{li2023ezinterviewer}. 
In this paper, we divide the person-job matching process into mock interview generation and two-sided evaluation in handshake protocol, jointly enhancing the performance of both modules to facilitate better person-job matching results.

\subsection{Capabilities of Role-Played LLMs}
With recent enhancements in model scales~\cite{chowdhery2022palm,ouyang2022training}, LLMs have shown impressive capabilities in various domains such as text generation~\cite{bubeck2023sparks,brown2020language}, ranking~\cite{ma2023large,chuang2023expand}, and evaluation/verification~\cite{shinn2023reflexion,madaan2023self}.
These powerful capabilities provide the opportunity to carefully design prompts to drive LLMs to play specific roles and complete specific tasks.
CharacterLLM gathers character profiles from Wikipedia and teaches LLMs to act as specific people~\cite{shao2023character}.
Wang et al. propose RoleLLM, a framework to benchmark, elicit, and enhance role-playing abilities in LLMs~\cite{wang2023rolellm}.
MetaGPT is a multi-agent framework for assigning different roles to GPTs to form a collaborative software entity for complex tasks~\cite{hong2023metagpt}. It is a specialized LLM-based multi-agent framework for collaborative software development.
CAMEL is an LLM-based communicative agent framework that demonstrates how role-playing can be used to enable chat agents to communicate with each other for task completion~\cite{li2023camel}.

However, traditional role-playing frameworks typically make a single LLM agent complete a specific sub-task, which is not in line with the requirements of interviewers and candidates. 
In this paper, we extend the role-playing framework to a multi-agent behavior collaboration paradigm. By assigning multiple behaviors for interviewers and candidates, they can generalize their capabilities in better simulation for job seeking and recruiting to promote better person-job matching.

\section{\model Framework}
In this section, we will formulate our definitions and notations for the person-job matching process, and then introduce the proposed framework \model in detail. 

\subsection{Problem Formulation}
The interactions between job seekers (\aka candidates) and job positions in the online recruitment data exhibit a two-sided one-to-many structure. Each candidate generally applies for multiple job positions, and each position, represented by a recruiter (\aka interviewer), is responsible for assessing numerous candidates.
Each candidate and job is accompanied by a descriptive text: a \textit{resume} documents the candidate's working skills and experiences, while a \textit{job description} outlines job requirements and responsibilities of the position.
In this paper, we divide the person-job matching process into two primary modules: (1) \textit{mock interview generation} and (2) \textit{two-sided evaluation in handshake protocol}.

Mock interview generation aims to simulate the multi-turn interview conversation between each interviewer $i_k$ and candidate $c_l$, where $k$ and $l$ are used as the indices of interviewers and candidates, respectively.
Given an interview dialogue context $U_n = \{q_1, a_1, \cdots, q_n, a_n\}$ containing $n$ historical question-answer pairs, the interviewer $i_k$ aims to pose a valuable and coherent question $q_{n+1}$ that aligns with the context $U_n$ and the candidate's resume. In response, the candidate $c_l$ is expected to generate an insightful answer $a_{n+1}$ that demonstrates their competencies.

Once each interview is completed, a two-sided evaluation is carried out by both interviewer $i_k$ and candidate $c_l$ to determine the suitability of the candidate for the job position.
Instead of traditional methods that solely rely on comparing the resume $r_l$ and job description $j_k$ to learn a matching function, we propose incorporating the mock interview dialogue history $U^{(i_k, c_l)}$ as a critical factor in person-job matching. Consequently, the final matching decision can be performed through a handshake protocol that considers the two-sided selection preferences, addressing the respective needs of both parties.
Furthermore, we incorporate reflection to summarize the essential interviewing experience and use the reflection memory $M^{(i_k)}$ and $M^{(c_l)}$ to dynamically modify the interview behaviors for both parties, improving their ability to refer to previous successful experience when facing new similar interviews.

\begin{figure*}
    \centering
    \includegraphics[width=\textwidth]{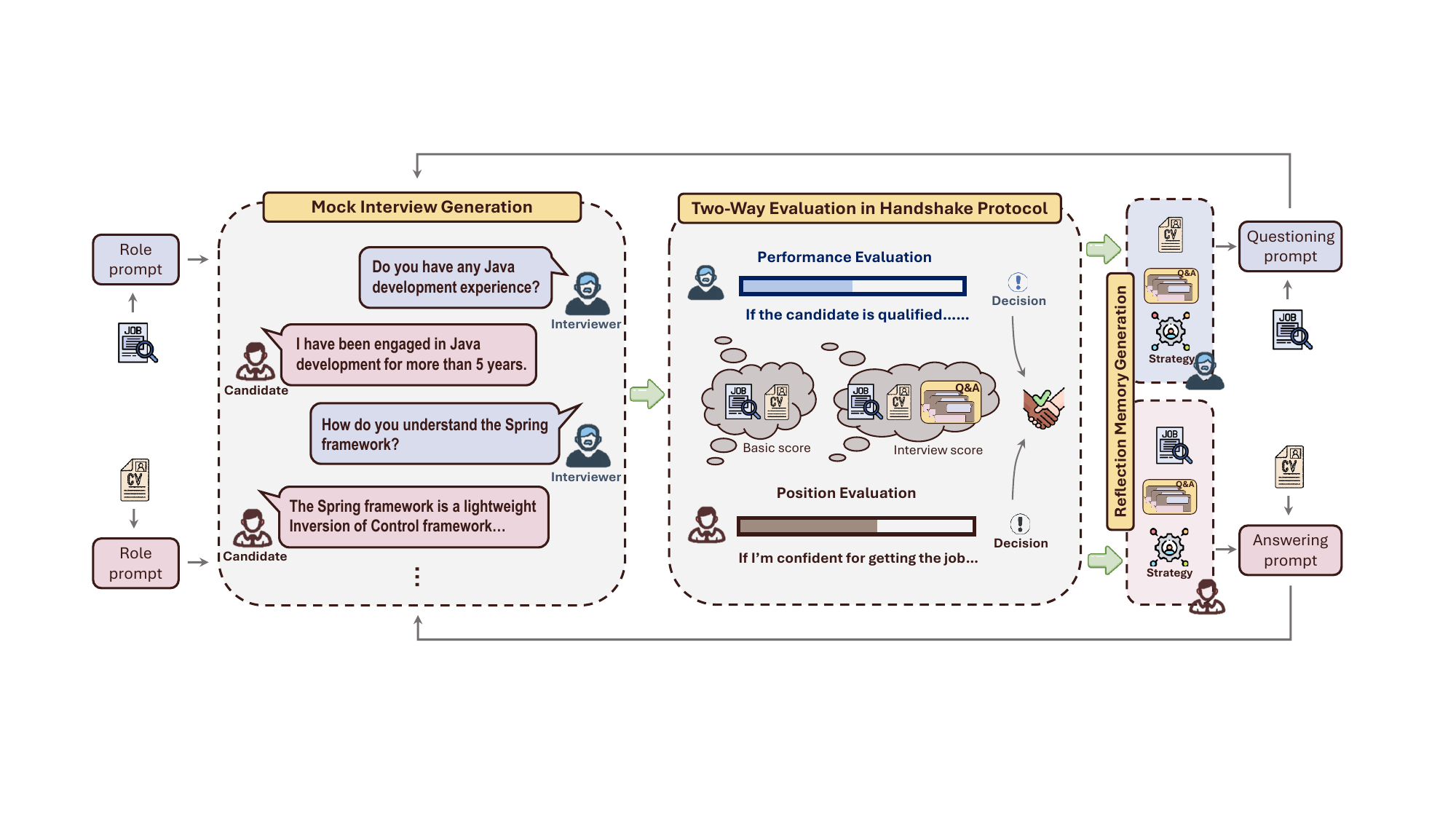}
    \caption{System overview of \model, which mainly consists of three modules: (1) Mock Interview Generation: The role-played interviewer and candidate form mock interviews through multi-turn conversations; (2) Two-Sided Evaluation in Handshake Protocol: Both parties evaluate each other based on the interview dialogue history, resume, and job description, and reach a matching consensus through a handshake protocol; (3) Reflection Memory Generation: Both roles save successfully matched cases to the reflection memory and apply them to modify questioning and answering strategies in subsequent interviews.}
    \label{fig:model}
\end{figure*}

\subsection{System Overview}
In this section, we introduce the system overview of the proposed \model, as illustrated in Figure~\ref{fig:model}.
First, we customize the different roles in the online recruitment data.
For each job description $j_k$, we instruct an LLM to comprehend the job requirements of $j_k$ and make it role-play an interviewer $i_k$ to communicate with job seekers and assess their performance.
Similarly, we prompt another LLM to play a candidate $c_l$ based on the key information (\eg skills and experiences) in each resume $r_l$ in response to the questions from the interviewer.
In the following subsections, we will use $f_*(\cdot)$ and $g_*(\cdot)$ as the functions for interviewer $i_k$ and candidate $c_l$, respectively. Thus, the role-playing process can be given by:
\begin{align}
    i_k & = f_{role}(j_k), \\
    c_l & = g_{role}(r_l).
\end{align}

To better formulate the person-job matching process, the proposed framework can be divided into two primary stages corresponding to the aforementioned modules: (1) the \textit{interview} stage for mock interview generation, and (2) the \textit{post-interview review} stage for two-sided evaluation in handshake protocol.
These two stages are performed alternately, making the job seeking and recruiting scenario a continuous process.
To enhance the ability of a single LLM role to perform multiple functions, we propose a multi-agent behavior collaboration paradigm to enable interviewers and candidates to effectively complete tasks at each stage of person-job matching.
For interviewers, their behaviors encompass \textit{interview question raising} during the interview stage, and \textit{interview performance evaluation} along with \textit{reflection memory generation} during the review stage.
As to job seekers, their behaviors include \textit{interview response generation} in the interview stage, and \textit{job position evaluation} as well as \textit{reflection memory generation} in the review stage.
Further details about these behaviors will be elaborated in subsequent subsections.

\subsection{Mock Interview Generation}
During the interview stage, interviewer $i_k$ and candidate $c_l$ engage in multiple turns of interaction in the conversation session.
The behavioral norms of both parties in this stage are crucial, as the quality of the interviewer's questions and the candidate's answers directly affects the usability of the mock interview. They also serve as the basis for the evaluation of each other by both parties in the review stage.
Therefore, to ensure the quality of the mock interview conversation, the goal of this module is to specify clear standards to guide the behaviors of interviewer $i_k$ and candidate $c_l$, controlling the generation of questions and responses.
This can be divided into interview question raising for the interviewer and interview response generation for the candidate.

\vspace{1mm}
\noindent \textbf{Interview question raising.}
Given an interview dialogue context $U_n = \{q_1, a_1, \cdots, q_n, a_n\}$ comprising $n$ historical question-answer interaction pairs, interviewer $i_k$ is responsible for raising a new question $q_{n+1}$ in the $n+1$ turn.
Due to the scarcity of expert resources, the LLM-played interviewers may not always prepare sufficiently valuable questions for systematically assessing candidates' competencies, particularly in the context of person-job matching.
To augment the professionalism of the interviewer, we establish well-defined standards for the question raising prompt $\theta_q$ concerning the posed question $q_{n+1}$: (1) coherence with the dialogue context $U_n$; (2) relevance to the job description $j_k$ and the candidate's resume $r_l$; and (3) diversity among previous questions $\{q_1, \cdots, q_n\}$ in $U_n$ to preclude repetition. In this way, we represent $f_{ques}$ as the question raising function, equipping all condition factors into $\theta_q$, and the newly raised question is thus formulated as:
\begin{equation}
    {q}_{n+1} = f_{ques}(U_n, r_l, j_k;\theta_q).
\end{equation}

\vspace{1mm}
\noindent \textbf{Interview response generation.}
As the behaviors of both roles are tightly linked through multi-turn dialogue interactions, the primary objective of candidate $c_l$ in the interview stage is to generate an accurate answer $a_{n+1}$ in response to the question $q_{n+1}$. Concurrently, the candidates' responses serve as a crucial factor in the subsequent two-sided evaluation and ultimate person-job matching.
We also clarify the response generation standards in the answering prompt $\theta_a$, specifying that the provided answer $a_{n+1}$ in response to $q_{n+1}$ should be coherent with the entire updated dialogue context $\widetilde{U}_n = \{q_1, a_1, \cdots, q_n, a_n, q_{n+1}\}$. Additionally, the candidate's insights and competencies described in the response must maintain relevance to the job description $j_k$ and the candidate's resume $r_l$.
The response generation process can be expressed by the function $g_{resp}$, as shown in the following equation:
\begin{equation}
    {a}_{n+1} = g_{resp}(\widetilde{U}_n, r_l, j_k;\theta_a).
\end{equation}

\subsection{Two-Sided Handshake Evaluation}
The post-interview review stage immediately commences once an interview is concluded. The primary objective of this stage is to determine whether the candidate and the job are matched based on their respective evaluations of each other. Unlike traditional person-job fitting methods computing matching scores between resumes and job descriptions~\cite{pjfnn,apjfnn}, the interview dialogue history provides alternative evidence for person-job matching decision-making.
We believe that modeling the two-sided preference selection of interviewers and candidates yields the most accurate person-job matching results.
Specifically, we integrate the interview dialogue history $U^{(i_k, c_l)}$ with the resume $r_l$ and job description $j_k$ to conduct a two-sided evaluation. This process can be divided into an interview performance evaluation for interviewer $i_k$ and a job position evaluation for candidate $c_l$. Then the proposed framework employs a handshake protocol based on the evaluation scores of both parties, forming a match only if both parties express mutual acceptance willingness.

\vspace{1mm}
\noindent \textbf{Interview performance evaluation.}
From the perspective of interviewer $i_k$, to evaluate each candidate $c_l$ more accurately and systematically, it can be considered to divide the evaluation into two scoring steps.
The interviewer first needs to give a basic fitting score $S_r^{(i_k \rightarrow c_l)}$ for descriptive texts by comparing whether the skills and experience in the resume $r_l$ are suitable for the job requirements in the job description $j_k$.
More importantly, the interviewer $i_k$ scores the candidate $c_l$'s performance $S_q^{(i_k \rightarrow c_l)}$ in the interview dialogue history $U^{(i_k, c_l)}$ to determine whether $c_l$'s response matches the story in $c_l$ and meets the job requirements in $j_k$.
Thus, the final evaluation score from $i_k$ to $c_l$ is related to a weighted average of $S_{r}^{(i_k \rightarrow c_l)}$ and $S_{q}^{(i_k \rightarrow c_l)}$, which can be formulated as:
\begin{align}
    S_{r}^{(i_k \rightarrow c_l)} & = f_{eval}(r_l, j_k), \\
    S_{q}^{(i_k \rightarrow c_l)} & = f_{eval}\left(U^{(i_k, c_l)}; r_l, j_k\right), \\
    E^{(i_k \rightarrow c_l)} & = \sigma \left( w_r\ S_{r}^{(i_k \rightarrow c_l)} + w_q\ S_{q}^{(i_k \rightarrow c_l)}\right),
\end{align}
where $\sigma(\cdot)$ denotes the sigmoid function.

\vspace{1mm}
\noindent \textbf{Job position evaluation.}
The evaluation behavior from the candidate's perspective is dual to that of the interviewer.
First, candidate $c_l$ needs to assign a basic fitting score $S_{j}^{(c_l\rightarrow i_k)}$ between his/her own resume and the requirements in job description $j_k$. Moreover, combining the interview dialogue history $U^{(i_k, c_l)}$, the candidate $c_l$ should further evaluate whether the position is suitable for him and score the opportunity of obtaining the offer for the position.
The final evaluation from $c_l$ to $i_k$ can also be viewed as a weighted average of $S_{j}^{(c_l\rightarrow i_k)}$ and $S_{a}^{(c_l\rightarrow i_k)}$, which is given by:
\begin{align}
    S_{j}^{(c_l\rightarrow i_k)} & = g_{eval}(r_l, j_k), \\
    S_{a}^{(c_l\rightarrow i_k)} & = g_{eval}\left(U^{(i_k, c_l)}; r_l, j_k\right), \\
    E^{(c_l\rightarrow i_k)} & = \sigma \left(w_j\ S_{j}^{(c_l\rightarrow i_k)}+ w_a\ S_{a}^{(c_l\rightarrow i_k)}\right).
\end{align}

\vspace{1mm}
\noindent \textbf{Person-job matching in hardshake protocol.}
Based on the comprehensive evaluation ensuring a more accurate and effective two-sided preference selection, the eventual decision of person-job making is performed by following a handshake protocol. Both the interviewer and the candidate have the right to choose whether to accept or reject each other, helping both parties better understand each other's needs and abilities, and promoting better person-job matching.
This handshake protocol can be formulated as:

\begin{equation}
    y(i_k,c_l) = \left\{
\begin{aligned}
    1, & \quad \text{if } E^{(i_k \rightarrow c_l)} = 1 \text{ and } E^{(c_l\rightarrow i_k)} = 1,\\
    0, & \quad \text{otherwise}.
\end{aligned}
\right.
\end{equation}

\subsection{Reflection Memory Generation}
The interaction during the interview stage provides a forward reference for the evaluation during the review stage. In order to better enable our framework to simulate real job seeking and recruiting scenarios, we consider feedback during the review stage to enhance the quality of the subsequent interview stage.
Specifically, in addition to the two-sided evaluation behaviors of job matching, we design \textit{reflection} behaviors for both parties during the review stage to summarize the experience in the interview and provide feedback to improve their behavior in subsequent interviews.
For both interviewers and candidates, we first initialize the natural language based questioning and answering strategies, which are denoted as $m_0^i$ and $m_0^c$, respectively.
Then, we consider leveraging the positive matching cases to continuously refine the strategies.
Taking interviewer $i_k$ as an example, if candidate $c_l$ achieves a two-sided matching with him/her, then their interview dialogue history $U^{(i_k,c_l)}$ is considered to be of relatively high quality and has guiding value for recruiting new candidates in future interviews.
Then, we use a reflection function $f_{ref}$ to refine the questioning strategy to $m^{(i_k\rightarrow c_l)}$, which can be more targeted to figure out the specific skills and experiences in the raised question.
Then, we incorporate $c_l$ into the $i_k$'s reflection memory $M^{(i_k)}$, including the tuple format: the resume $r_l$, the interview dialogue history $U^{(i_k,c_l)}$, and the refined questioning strategy $m^{(i_k\rightarrow c_l)}$.
Similar to each candidate $c_l$, the answering strategy can be refined to $m^{(c_l\rightarrow i_k)}$ through $g_{ref}$.

\begin{equation}
m^{(i_k\rightarrow c_l)} = \left\{
    \begin{aligned} &
    f_{ref}\left(r_l, U^{(i_k, c_l)}\right), & \quad \text{if } y(i_k,c_l) = 1. \\
    & m_0^{i}, & \quad \text{otherwise}.
\end{aligned}
\right.
\end{equation}
\begin{equation}
    M^{(i_k)} \leftarrow M^{(i_k)}\bigcup\big\{c_l|r_l, U^{(i_k, c_l)}, m^{(i_k\rightarrow c_l)}\big\}.
\end{equation}

\begin{equation}
m^{(c_l\rightarrow i_k)} = \left\{
    \begin{aligned} &
    g_{ref}\left(j_k, U^{(i_k, c_l)}\right), & \quad \text{if } y(i_k,c_l) = 1. \\
    & m_0^{c}, & \quad \text{otherwise}.
\end{aligned}
\right.
\end{equation}
\begin{equation}
    M^{(c_l)} \leftarrow M^{(c_l)}\bigcup\big\{i_k|j_k, U^{(i_k, c_l)}, m^{(c_l\rightarrow i_k)}\big\}.
\end{equation}

\subsection{Strategy Modification with Feedback}
In order to further link the two stages of interview and review, we consider using the feedback from the review stage to help improve the quality of subsequent interviews.
Specifically, the feedback is the memory formed by reflection. The specific optimization method is to modify the strategies for interview questions and responses to make them more targeted.
The initial questioning strategy applies to all candidates, but after modification, the interviewer will ask questions more specifically based on the skills and experience of different interviewers. The same goes for job seekers. The prompts they reply to are initially short and general, but as the number of interviews increases, the interview experience is accumulated through the reflection process.
The strategy modification of questioning lies in two aspects: (1) Based on the summary content formed by reflection, the strategy for questioning is improved, so that the questions asked in different candidates and different rounds are more targeted; (2) Extracting similar interviews from memory Information allows the interviewer to find job seekers with similar job skills or experience that have been interviewed before in the memory, and use the questions they asked as a reference when asking new candidates.

\begin{align}
    {\theta}_q & \leftarrow f_{mod} (\theta_q, M^{(i_k)}_{c'}, r'), \\
    {\theta}_a & \leftarrow f_{mod} (\theta_a, M^{(c_l)}_{i'}, j'),
\end{align}

\begin{table*}
  \centering
  \caption{Performance comparison of two-sided person-job matching on the PJF-Base dataset. Bold numbers indicate that we accept the hypothesis of model improvement relative to the best baseline at a significance test level of 0.01.}
    \begin{tabular}{@{}l|cccc|cccc|ccc@{}}
    \toprule
    \textbf{Direction} & \multicolumn{4}{c|}{\textbf{Candidates}} & \multicolumn{4}{c|}{\textbf{Jobs}} & \multicolumn{3}{c}{\textbf{Both Sides}} \\
    \midrule
    \textbf{Method} & \textbf{NDCG@5} & \textbf{R@5} & \textbf{P@5} & \textbf{MRR@5} & \textbf{NDCG@5} & \textbf{R@5} & \textbf{P@5} & \textbf{MRR@5} & \textbf{Precision} & \textbf{Recall} & \textbf{F1} \\
    \midrule
    NCF~\cite{ncf}   & 0.296  & 0.356  & 0.207  & 0.401  & 0.273  & 0.329  & 0.193  & 0.375  & 0.508  & 0.516  & 0.425  \\
    APJFNN~\cite{apjfnn} & 0.269  & 0.324  & 0.186  & 0.378  & 0.258  & 0.314  & 0.188  & 0.348  & 0.501  & 0.502  & 0.431  \\
    PJFNN~\cite{pjfnn} & 0.275  & 0.332  & 0.192  & 0.374  & 0.260  & 0.318  & 0.187  & 0.356  & 0.503  & 0.505  & 0.453  \\
    IPJF~\cite{ipjf}  & 0.353  & 0.427  & 0.252  & 0.440  & 0.340  & 0.420  & 0.250  & 0.441  & 0.549  & 0.596  & 0.477  \\
    \midrule
    \model w/o MI & 0.218  & 0.297  & 0.167  & 0.236  & 0.199  & 0.258  & 0.146  & 0.223  & 0.591  & 0.622  & 0.578  \\
    \model w/o SU & 0.366  & 0.450  & 0.261  & 0.422  & 0.376  & 0.463  & 0.271  & 0.432  & 0.597  & 0.659  & 0.583  \\
    \model w/o Ref & 0.372  & 0.453  & 0.262  & 0.430  & 0.380  & 0.470  & 0.277  & 0.434  & 0.590  & 0.651  & 0.560  \\
    \textbf{\model} & \textbf{0.395} & \textbf{0.479} & \textbf{0.279} & \textbf{0.443} & \textbf{0.413} & \textbf{0.494} & \textbf{0.296} & \textbf{0.475} & \textbf{0.611} & \textbf{0.685} & \textbf{0.586} \\
    \bottomrule
    \end{tabular}
  \label{tab:main}
\end{table*}

\section{Experiments}
\subsection{Setup}
\noindent \textbf{Datasets.}
In this paper, we experiment with the online recruitment datasets sourced from ``Boss Zhipin''\footnote{https://www.zhipin.com}, China's largest online recruiting platform.  
We collect two datasets from real online platform logs for evaluation, \textbf{PJF-Base} and \textbf{PJF-Large}, with no period overlap between them.
PJF-Base is carefully curated, retaining complete and high-quality resumes and job descriptions from 106 days of online data.
It contains a total of 1,992 resumes and 1,968 job descriptions. 
To evaluate the model performance on more diverse and larger-scale data, PJF-Large has been collected from new logs over the past six months. It covers about 10k candidates and 16k job positions in total.
To protect candidate privacy, all textual documents are anonymized by removing personal identification information.

\vspace{1mm}
\noindent \textbf{Evaluation metrics.}
Following the previous studies~\cite{zhao2022revisiting}, we employ four widely used top-$k$ recommendation metrics to test the effectiveness of person-job matching from both candidate and job perspectives: \textbf{Recall (R@k)}, \textbf{Precision (P@k)}, \textbf{Normalized Discounted Cumulative Gain (NDCG@k)}, and \textbf{Mean Reciprocal Rank (MRR@k)}. 
To further evaluate two-sided matching performance, we use Macro Precision, Recall, and F1 scores to measure the accuracy of successful matching between both parties.

To evaluate the quality of mock interview generation, we use \textbf{BLEU (B@n)} for n-gram overlap between generated utterances and candidates' resumes.

Moreover, we conduct human evaluations to comprehensively assess the mock interview quality, using the following metrics: (1) \textbf{Coherence}: The extent to which the generated interview questions/answers are coherent with the dialogue context; (2) \textbf{Relevance}: The degree to which the generated interviews are relevant to the candidate's resume; (3) \textbf{Diversity}: The variation in the generated interviews compared to previous dialogue history.

\vspace{1mm}
\noindent \textbf{Comparisons.}
To evaluate the effectiveness of our framework for person-job matching, we compare it with the following methods: (1) \textbf{NCF}~\cite{ncf} leverages collaborative filtering to model the interaction between jobs and candidates; (2) \textbf{PJFNN}~\cite{pjfnn} employs convolutional neural networks to encode resumes and job descriptions separately and computes their cosine similarity scores; (3) \textbf{APJFNN}~\cite{apjfnn} leverages semantic representations for both resumes and job descriptions based on sequential networks and hierarchical attention; (4) \textbf{IPJF}~\cite{ipjf} uses the multi-level interactions as supervision signals to indicate the preferences for candidates and jobs.

For assessing the quality of mock interview generation, we fine-tune classical specialized models: \textbf{GPT-2}~\cite{gpt2} and \textbf{T5-base}~\cite{t5} in the \texttt{TecentPretrain}~\cite{zhao2022tencentpretrain} toolkit on our training data for comparison. Inspired by knowledge-grounded dialogue generation methods, we use \textbf{K2R}~\cite{adolphs2021reason} to first generate a knowledge sentence based on the context and then produce a final response.
We also include a baseline named \textbf{LLM-Direct}, which uses the same backbone as our \model but directly simulates the interview conversations without role-playing and strategy updating processes.

\begin{table*}
  \centering
  \caption{Performance comparison of two-sided person-job matching on the PJF-Large dataset. Bold numbers indicate that we accept the hypothesis of model improvement relative to the best baseline at a significance test level of 0.01.}
    \begin{tabular}{@{}l|cccc|cccc|ccc@{}}
    \toprule
    \textbf{Direction} & \multicolumn{4}{c|}{\textbf{Candidates}} & \multicolumn{4}{c|}{\textbf{Jobs}} & \multicolumn{3}{c}{\textbf{Both Sides}} \\
    \midrule
    \textbf{Method} & \textbf{NDCG@5} & \textbf{R@5} & \textbf{P@5} & \textbf{MRR@5} & \textbf{NDCG@5} & \textbf{R@5} & \textbf{P@5} & \textbf{MRR@5} & \textbf{Precision} & \textbf{Recall} & \textbf{F1} \\
    \midrule
NCF~\cite{ncf}   & 0.351 & 0.372 & 0.257 & 0.399 & 0.302 & 0.35  & 0.278 & 0.394 & 0.511 & 0.512 & 0.44 \\
APJFNN~\cite{apjfnn} & 0.359 & 0.356 & 0.264 & 0.378 & 0.285 & 0.361 & 0.291 & 0.37  & 0.496 & 0.502 & 0.423 \\
    PJFNN~\cite{pjfnn} & 0.367 & 0.375 & 0.261 & 0.371 & 0.277 & 0.349 & 0.284 & 0.368 & 0.507 & 0.509 & 0.436 \\
    IPJF~\cite{ipjf}  & 0.372 & 0.418 & 0.282 & 0.416 & 0.325 & 0.387 & 0.306 & 0.422 & 0.529 & 0.533 & 0.467 \\
    \midrule
    \model w/o MI    & 0.324 & 0.343 & 0.227 & 0.358 & 0.284 & 0.325 & 0.258 & 0.353 & 0.522 & 0.531 & 0.503 \\
    \model w/o SU    & 0.522 & 0.498 & 0.301 & 0.476 & 0.546 & 0.526 & 0.355 & 0.497 & 0.612 & 0.624 & 0.574 \\
    \model w/o Ref    & 0.534 & 0.506 & 0.306 & 0.480  & 0.552 & 0.539 & 0.363 & 0.501 & 0.627 & 0.631 & 0.588 \\
    \textbf{MockLLM} & \textbf{0.566} & \textbf{0.523} & \textbf{0.367} & \textbf{0.515} & \textbf{0.574} & \textbf{0.544} & \textbf{0.372} & \textbf{0.521} & \textbf{0.643} & \textbf{0.658} & \textbf{0.601} \\
    \bottomrule
    \end{tabular}
  \label{tab:large}
\end{table*}

\begin{table*}
  \centering
  \caption{Automatic and human evaluation results of mock interview generation.}
    \begin{tabular}{@{}l|cccc|cccc|ccc@{}}
    \toprule
    \multirow{2}[4]{*}{\textbf{Method}} & \multicolumn{4}{c|}{\textbf{Questions}} & \multicolumn{4}{c|}{\textbf{Answers}} & \multicolumn{3}{c}{\textbf{Human Eval}} \\
\cmidrule{2-12}          & {\textbf{B@1}} & {\textbf{B@2}} & {\textbf{B@3}} & {\textbf{B@4}} & {\textbf{B@1}} & {\textbf{B@2}} & {\textbf{B@3}} & {\textbf{B@4}} & \textbf{Coherence} & \textbf{Relevance} & \textbf{Diversity} \\
    \midrule
    GPT-2 & 13.74  & 8.76  & 5.45  & 3.40  & 2.53  & 1.23  & 0.58  & 0.27  & 2.26  & 2.05  & 2.08  \\
    T5 & 17.70  & 13.53  & 10.58  & 8.52  & 2.52  & 1.48  & 0.89  & 0.59  & 3.21  & 3.34  & 3.15  \\
    GPT-2+K2R  & 14.24  & 8.50  & 4.74  & 2.62  & 2.72  & 1.33  & 0.60  & 0.27  & 2.34  & 1.99  & 2.16  \\
    T5+K2R & 18.36  & 13.35  & 9.92  & 7.68  & 2.96  & 1.70  & 0.99  & 0.62  & 3.45  & 3.37  & 3.23  \\
    LLM-Direct & 19.22 & 13.21 & 10.01 & 7.92 & 9.77 & 4.32 & 2.29 & 1.93 & 3.74 & 3.65 & 3.43 \\
    \midrule
    \textbf{\model} & \textbf{24.23} & \textbf{16.75} & \textbf{12.14} & \textbf{9.20} & \textbf{16.48} & \textbf{9.87} & \textbf{6.36} & \textbf{4.53} & \textbf{4.12} & \textbf{4.03} & \textbf{4.20} \\
    \bottomrule
    \end{tabular}
  \label{tab:interview}
\end{table*}

\subsection{Overall Performance}
We conduct comprehensive comparative experiments on two benchmark datasets, PJF-Based and PJF-Large, to evaluate the effectiveness of our proposed framework in person-job matching tasks.
The results on PJF-Based and PJF-Large are shown in Tables~\ref{tab:main} and \ref{tab:large}.
Among existing methods, content-based approaches such as APJFNN and PJFNN underperform relative to collaborative filtering-based models like NCF. This performance gap can be attributed to the limitations of content-based models, which heavily rely on textual representations of resumes and job descriptions. In real-world online recruitment data, such textual inputs often exhibit considerable variability and lack consistent structural formatting.
In contrast, NCF leverages similar matching cases to infer the compatibility for new cases, resulting in slightly better performance. By learning latent factors based on similar candidate-job pairs, it achieves moderately better results, though still limited in handling complex textual semantics.
The hybrid-based IPJF model can further enhance performance by combining textual document encodings with information from similar historical matches. This integration helps IPJF balance generalization and specificity, yielding improved matching accuracy across evaluation metrics.

Our proposed framework, \model, consistently outperforms all baselines in terms of matching accuracy for candidates, jobs, and both sides. The advantage becomes even more evident when the experiment is scaled up to the larger PJF-Large dataset, where \model maintains superior performance across all matching metrics.
These results highlight the robustness and scalability of our approach. Unlike traditional models, \model captures both semantic and interactive dimensions of the recruitment process.
\model benefits from modeling textual documents into role-playing LLMs, using reflection memory to consider successful matching experiences, and leveraging handshake evaluation to improve matching satisfaction for both candidates and interviewers.

\subsection{Ablation Study}
To verify the effectiveness of \model's components, we conduct an ablation study with the following variants: (1) \textbf{\model w/o MI} eliminates mock interview generation, relying solely on resumes and job descriptions for evaluation; (2) \textbf{\model w/o Ref} removes the reflection behaviors and strategy modification, retaining only the forward process of person-job matching; (3) \textbf{\model w/o SU} omits strategy updates and consistently uses the initial strategies $m_0^i$ and $m_0^c$ for all mock interviews.
The ablation results shown in Table~\ref{tab:main} show that \model w/o MI performs the worst, underscoring the importance of integrating mock interview dialogues. The performance gap between \model w/o Ref and the full model demonstrates the effectiveness of the reflection memory generation in refining interviewer and candidate behaviors. Finally, \model w/o SU fails to generate targeted and valuable questions or answers, confirming the necessity of strategy updates within the framework.

\subsection{Impact of Mock Interview Generation}
As a key module of our framework, the quality of mock interview generation directly influences the matching evaluation criteria and needs to be carefully evaluated.

\vspace{1mm}
\noindent \textbf{Automatic evaluation.}
As shown in the left part of Table ~\ref{tab:interview}, our \model outperforms specialized dialogue models in mock interview generation. Although K2R uses generated outcomes as intermediate references to slightly improve the document relevance, it remains inferior to \model. 
Additionally, LLM-Direct also performs less effectively than \model. These results suggest that the superiority of our framework stems not only from the use of LLMs but also from the thoughtful design of essential modules such as role-playing and strategy updating.

\vspace{1mm} 
\noindent \textbf{Human evaluation.}
We also employ three well-educated volunteers to evaluate the quality of generated interview questions and responses. This double-blind evaluation ensures annotators are unaware of which method generated which output. We randomly sample 100 cases from the test set and ask each annotator to rate the questions and responses generated by each method based on the metrics: \textbf{Coherence}, \textbf{Relevance}, and \textbf{Diversity}.
The human evaluation results are listed on the right side of Table~\ref{tab:interview}.
Compared to baseline methods, our proposed \model can generate more coherent questions and responses to align with the interview context.
The higher Relevance score indicates that the questions raised by \model are more relevant to the candidate's resume, making the interview topics accurate and valuable.
Moreover, the highest Diversity means that \model can generate varied questions from new directions based on existing questions, reducing repetition and enhancing the quality of mock interviews.

\subsection{Scalability Analysis}
To demonstrate our framework's broader applicability beyond the specific candidates and jobs in PJF-Base, we conduct a scalability analysis on the PJF-Large dataset.
We compare our framework with the strong baseline IPJF~\cite{ipjf} across varying data sizes to evaluate the impact on their matching performance.
By gradually increasing the number of candidates to be evaluated on PJF-Large, we examine how the proportion of successful matches between candidates and suitable positions changes with data size.

\begin{figure}[t]
    \centering
    \includegraphics[width=0.88\linewidth]{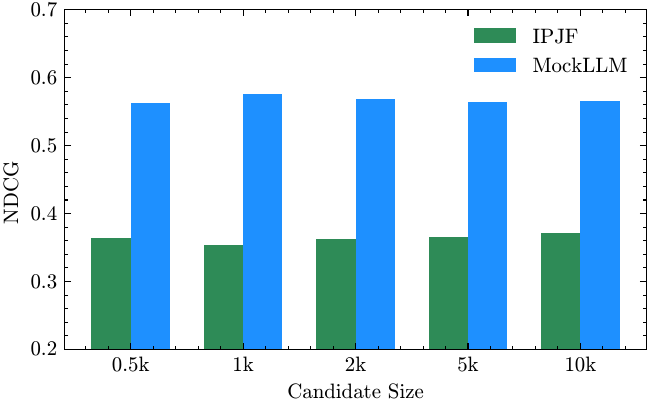}
    \caption{The effect of varying candidate sizes.}
    \label{fig:scale}
\end{figure}

The results in Figure~\ref{fig:scale} show that our framework maintains consistent matching performance across different data scales, with no significant fluctuations. 
Notably, the NDCG score of \model always outperforms the baseline IPJF by about 0.2 points. These findings indicate that our proposed method not only scales effectively with larger datasets but also maintains high matching accuracy, making it more robust and useful for real-world applications under varying data sizes.

\subsection{Domain Adaption Analysis}
Person-job matching tasks require adaptability to varying recruitment requirements, interview assessment focuses, and response skills across different domains.
Therefore, we review all job positions in the collected data and summarize them into four categories based on the job description requirements: \textit{Testing}, \textit{Operation}, \textit{Development}, and \textit{Others}. 

\begin{figure}[t]
    \centering
    \includegraphics[width=0.88\linewidth]{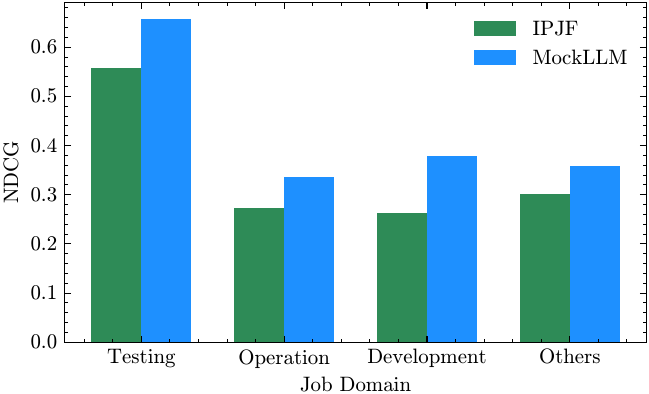}
    \caption{The effect of different job domains.}
    \label{fig:dom}
\end{figure}

The comparison results in Figure~\ref{fig:dom} highlight performance variations across these domains. While IPJF performs decently in Testing, its performance drops significantly in other domains, indicating a strong dependency on the domain. In contrast, \model shows consistent superiority over IPJF across all job categories, underscoring the considerable adaptability of our framework to diverse domains. 
\model can leverage interview questions from previously matched candidates in the interviewer's reflection memory, allowing for continuous refinement of questioning strategies.
Thus, as interviewers accumulate more experience in their specific domains, their questioning strategies become increasingly specialized, enhancing the framework's ability to adapt to specific job domains.

\subsection{Fairness Analysis}
Ensuring fairness in person-job matching systems is critical, particularly in sensitive domains such as recruitment, where algorithmic decisions can significantly impact individuals’ career opportunities. To protect user privacy, we anonymize all resumes prior to their integration into our framework, removing any personally identifiable information that could influence the model's behavior in unintended ways.

To further investigate potential fairness concerns, we conduct an analysis based on the sensitive attribute gender. Specifically, we divide the test set into \textit{Male} and \textit{Female} subsets to compare the performance of our \model with the strong baseline IPJF~\cite{ipjf}.
The results indicate disparities in person-job matching across genders. However, our \model demonstrates reduced bias compared to the baseline, suggesting that our framework mitigates fairness issues to some extent.

\begin{table}[t]
    \centering
    \caption{Comparison of IPJF and MockLLM by gender.}
    \resizebox{\linewidth}{!}{
    \begin{tabular}{llcccc}
        \toprule
        \textbf{Gender} & \textbf{Method} & \textbf{NDCG} & \textbf{Recall} & \textbf{Precision} & \textbf{MRR} \\
        \midrule
        \multirow{2}{*}{Male} & IPJF & 0.391 & 0.444 & 0.212 & 0.454 \\
                              & \textbf{MockLLM} & \textbf{0.447} & \textbf{0.512} & \textbf{0.292} & \textbf{0.529} \\
        \midrule
        \multirow{2}{*}{Female} & IPJF & 0.217 & 0.221 & 0.192 & 0.315 \\
                                & \textbf{MockLLM} & \textbf{0.385} & \textbf{0.468} & \textbf{0.273} & \textbf{0.440} \\
        \bottomrule
    \end{tabular}
    }
    \label{tab:comparison}
\end{table}

\begin{figure*}
    \centering
    \includegraphics[width=\textwidth]{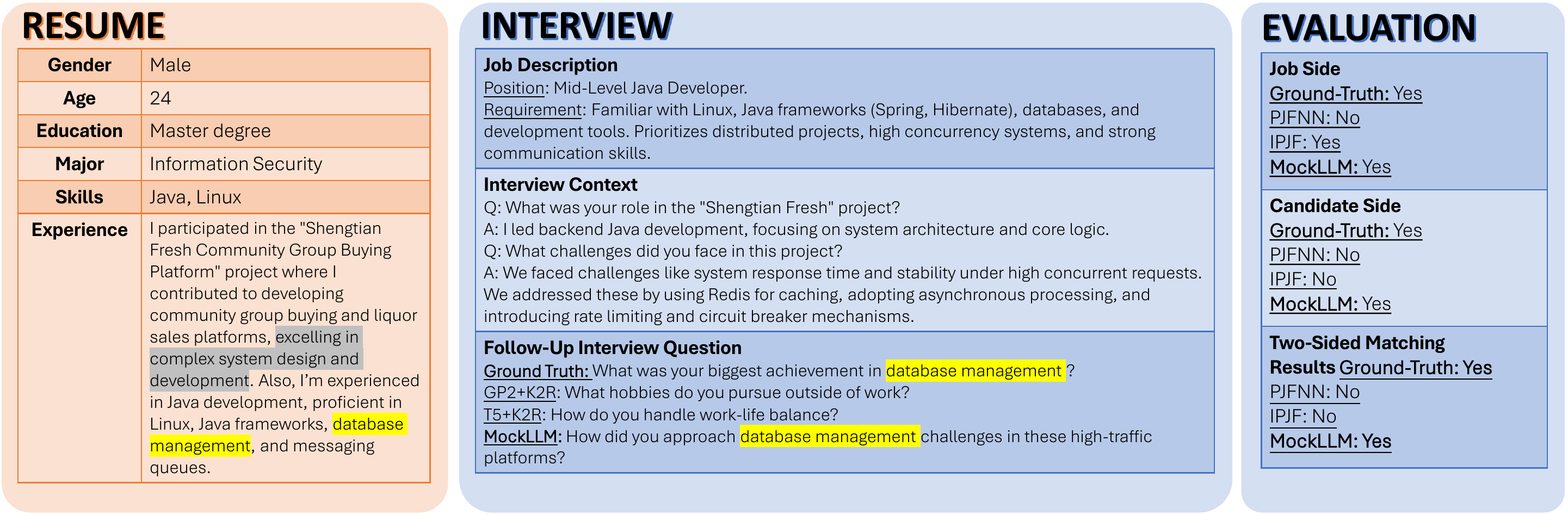}
    \caption{An example of mock interview generation and two-sided evaluation shows the superiority of \model, which can generate higher-quality mock interviews and achieve better two-sided person-job matching decisions.}
    \label{fig:case}
\end{figure*}

\begin{figure}
    \centering
    \includegraphics[width=\linewidth]{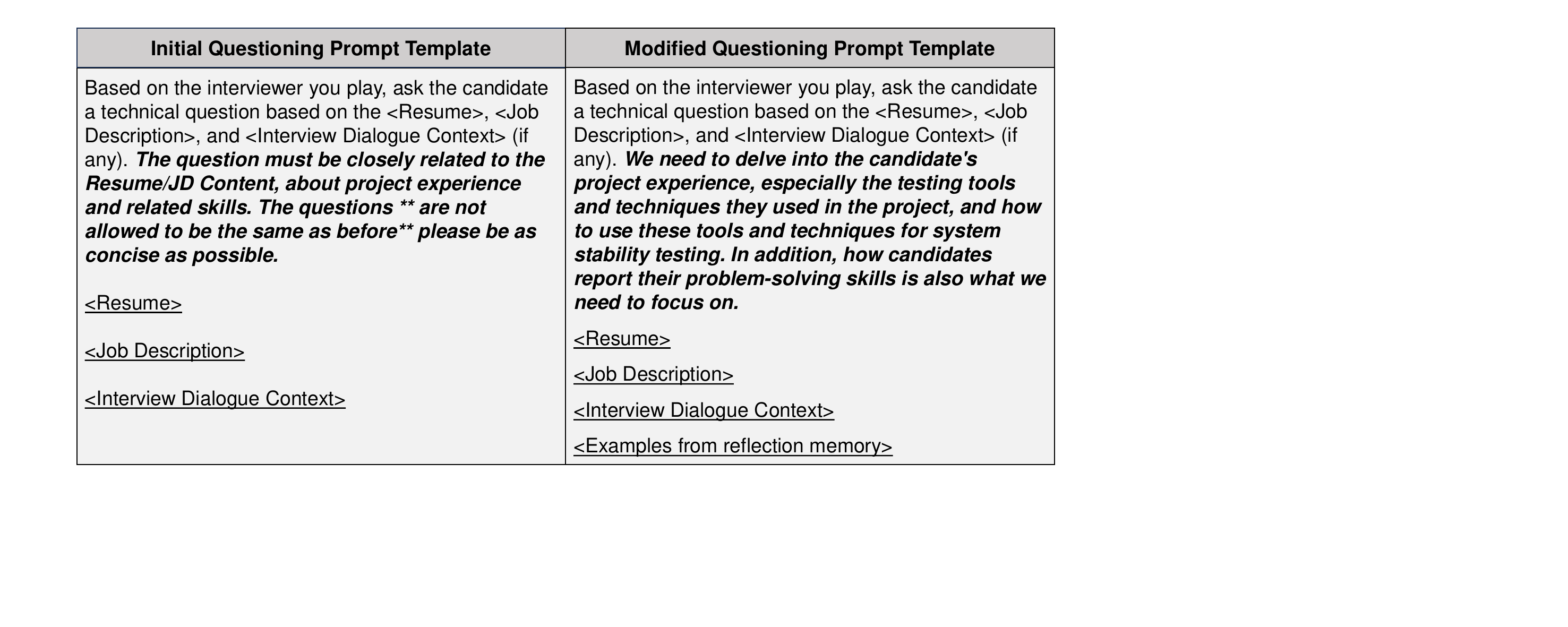}
    \caption{Comparison results before and after questioning strategy modification.}
    \label{fig:caseprompt}
\end{figure}

\subsection{Qualitative Analysis}

\vspace{1mm}
\noindent \textbf{Case study.}
We present a person-job matching example for comparison with our proposed \model and baseline models in Figure~\ref{fig:case}.
From the given resume, we observe that the candidate is proficient in Java and Linux, and has experience in database management projects. The interview questions simulated by the interviewer in \model not only match the candidate's background but also enable targeted conversations revolved around their key skills and experiences. In contrast, baseline models fail to raise valuable questions that are sufficiently connected to the resume in multi-turn conversations.
As to the evaluation process, only our method reaches the correct matching results through two-sided evaluation in handshake protocol, where baseline methods fail to make an accurate matching decision as they cannot incorporate mock interview history into the evaluation process.

\vspace{1mm}
\noindent \textbf{Analysis of strategy modification.}
Regarding our proposed strategy modification technique, we provide specific results before and after questioning strategy modification and give more elaborations for qualitative comparative analysis. Figure~\ref{fig:caseprompt} shows that the initial questioning strategy without modification is a general instruction for all candidates.
Compared to the initial strategy, the modified strategy expands more questioning requirements and insights in detail and incorporates relevant cases from reflection memory to draw on prior experience to help raise new valuable questions.

\subsection{Online Evaluation}
To assess the assistance of the proposed framework on online recruitment, we establish a simulated online environment identical to the real-world platform and conduct internal testing.
A group of real beta users is invited to submit resumes and interact with the system in realistic scenarios.
We designate the original person-job matching pipeline used on the platform as \textbf{Pipeline}. As a comparison, we incorporate the \model framework into the system as an enhanced version named \textbf{Pipeline+MockLLM}, which utilizes role-playing LLMs for mock interview generation and two-sided evaluation.
We conduct an A/B test to compare their performance. The results of two-sided matching metrics in Table~\ref{tab:online} indicate a significant improvement (all $p$-values<0.01) with the adoption of our framework, revealing that \model enhances the online system's ability to find more satisfactory job matches for candidates. This demonstrates the potential gains of \model for better satisfaction for job seekers and recruiters.
The system has been integrated into a simulated real-world pipeline, ensuring compliance with operational constraints and its deployment readiness. The design and results of this testing phase reflect a clear pathway to broader adoption.
Currently, the inference time per case after integrating our \model is about 0.39s, an acceptable increase given the performance gains.
This testing phase has provided valuable insights for further optimization and sets a clear path toward full deployment.

\begin{table}
    \centering
    \caption{The online evaluation results. Bold numbers indicate significant gains with all $p$-values lower than 0.01.}
    
    \begin{tabular}{lccc}
\toprule
\textbf{Metric} & \textbf{Pipeline} & \textbf{Pipeline+MockLLM} & \textbf{$p$-value} \\
\midrule
{Recall} & 0.30 & \textbf{0.60} & 3e-14 \\
{Precision} & 0.40 & \textbf{0.45} & 0.008 \\
{F1 Score} & 0.34 & \textbf{0.51} & 2e-10 \\
{Latency} & 0.003s & 0.39s & - \\
\bottomrule
\end{tabular}
    \label{tab:online}
\end{table}

\section{Conclusion}
This paper presents \model, a novel framework leveraging LLM role-playing capabilities to form mock interview conversations, enriching evidence for more effective person-job matching. By incorporating multi-agent behavior collaboration, reflection memory generation, and dynamic strategy modification, MockLLM continuously refines the behaviors of interviewers and candidates. The framework showcases strong potential for delivering scalable and adaptable solutions that advance recruitment processes beyond the limitations of traditional methods.

\begin{acks}
This work was supported by the Beijing
Outstanding Young Scientist Program NO. BJJWZYJH012019100020098, and Intelligent Social Governance Platform, Major Innovation \& Planning Interdisciplinary Platform for the ``Double-First Class'' Initiative, Renmin University of China, the Fundamental Research Funds for the Central Universities, and Public Computing Cloud, Renmin University of China, the fund for building world-class universities (disciplines) of Renmin University of China.
\end{acks}

\newpage
\bibliographystyle{ACM-Reference-Format}
\balance
\bibliography{mockllm}


\begin{thebibliography}{39}

%%% the Web address.

\ifx \showCODEN    \undefined \def \showCODEN     #1{\unskip}     \fi
\ifx \showISBNx    \undefined \def \showISBNx     #1{\unskip}     \fi
\ifx \showISBNxiii \undefined \def \showISBNxiii  #1{\unskip}     \fi
\ifx \showISSN     \undefined \def \showISSN      #1{\unskip}     \fi
\ifx \showLCCN     \undefined \def \showLCCN      #1{\unskip}     \fi
\ifx \shownote     \undefined \def \shownote      #1{#1}          \fi
\ifx \showarticletitle \undefined \def \showarticletitle #1{#1}   \fi
\ifx \showURL      \undefined \def \showURL       {\relax}        \fi
% invisible to TeX
\providecommand\bibfield[2]{#2}
\providecommand\bibinfo[2]{#2}
\providecommand\natexlab[1]{#1}
\providecommand\showeprint[2][]{arXiv:#2}

\bibitem[Adolphs et~al\mbox{.}(2021)]%
        {adolphs2021reason}
\bibfield{author}{\bibinfo{person}{Leonard Adolphs}, \bibinfo{person}{Kurt Shuster}, \bibinfo{person}{Jack Urbanek}, \bibinfo{person}{Arthur Szlam}, {and} \bibinfo{person}{Jason Weston}.} \bibinfo{year}{2021}\natexlab{}.
\newblock \showarticletitle{Reason first, then respond: Modular generation for knowledge-infused dialogue}.
\newblock \bibinfo{journal}{\emph{arXiv preprint arXiv:2111.05204}} (\bibinfo{year}{2021}).
\newblock


\bibitem[Brown et~al\mbox{.}(2020)]%
        {brown2020language}
\bibfield{author}{\bibinfo{person}{Tom Brown}, \bibinfo{person}{Benjamin Mann}, \bibinfo{person}{Nick Ryder}, \bibinfo{person}{Melanie Subbiah}, \bibinfo{person}{Jared~D Kaplan}, \bibinfo{person}{Prafulla Dhariwal}, \bibinfo{person}{Arvind Neelakantan}, \bibinfo{person}{Pranav Shyam}, \bibinfo{person}{Girish Sastry}, \bibinfo{person}{Amanda Askell}, {et~al\mbox{.}}} \bibinfo{year}{2020}\natexlab{}.
\newblock \showarticletitle{Language models are few-shot learners}.
\newblock \bibinfo{journal}{\emph{Advances in neural information processing systems}}  \bibinfo{volume}{33} (\bibinfo{year}{2020}), \bibinfo{pages}{1877--1901}.
\newblock


\bibitem[Bubeck et~al\mbox{.}(2023)]%
        {bubeck2023sparks}
\bibfield{author}{\bibinfo{person}{S{\'e}bastien Bubeck}, \bibinfo{person}{Varun Chandrasekaran}, \bibinfo{person}{Ronen Eldan}, \bibinfo{person}{Johannes Gehrke}, \bibinfo{person}{Eric Horvitz}, \bibinfo{person}{Ece Kamar}, \bibinfo{person}{Peter Lee}, \bibinfo{person}{Yin~Tat Lee}, \bibinfo{person}{Yuanzhi Li}, \bibinfo{person}{Scott Lundberg}, {et~al\mbox{.}}} \bibinfo{year}{2023}\natexlab{}.
\newblock \showarticletitle{Sparks of artificial general intelligence: Early experiments with gpt-4}.
\newblock \bibinfo{journal}{\emph{arXiv preprint arXiv:2303.12712}} (\bibinfo{year}{2023}).
\newblock


\bibitem[Chen et~al\mbox{.}(2023a)]%
        {chen2023autoagents}
\bibfield{author}{\bibinfo{person}{Guangyao Chen}, \bibinfo{person}{Siwei Dong}, \bibinfo{person}{Yu Shu}, \bibinfo{person}{Ge Zhang}, \bibinfo{person}{Jaward Sesay}, \bibinfo{person}{B{\"o}rje~F Karlsson}, \bibinfo{person}{Jie Fu}, {and} \bibinfo{person}{Yemin Shi}.} \bibinfo{year}{2023}\natexlab{a}.
\newblock \showarticletitle{Autoagents: A framework for automatic agent generation}.
\newblock \bibinfo{journal}{\emph{arXiv preprint arXiv:2309.17288}} (\bibinfo{year}{2023}).
\newblock


\bibitem[Chen et~al\mbox{.}(2023b)]%
        {chen2023agentverse}
\bibfield{author}{\bibinfo{person}{Weize Chen}, \bibinfo{person}{Yusheng Su}, \bibinfo{person}{Jingwei Zuo}, \bibinfo{person}{Cheng Yang}, \bibinfo{person}{Chenfei Yuan}, \bibinfo{person}{Chen Qian}, \bibinfo{person}{Chi-Min Chan}, \bibinfo{person}{Yujia Qin}, \bibinfo{person}{Yaxi Lu}, \bibinfo{person}{Ruobing Xie}, {et~al\mbox{.}}} \bibinfo{year}{2023}\natexlab{b}.
\newblock \showarticletitle{Agentverse: Facilitating multi-agent collaboration and exploring emergent behaviors in agents}.
\newblock \bibinfo{journal}{\emph{arXiv preprint arXiv:2308.10848}} (\bibinfo{year}{2023}).
\newblock


\bibitem[Chowdhery et~al\mbox{.}(2022)]%
        {chowdhery2022palm}
\bibfield{author}{\bibinfo{person}{Aakanksha Chowdhery}, \bibinfo{person}{Sharan Narang}, \bibinfo{person}{Jacob Devlin}, \bibinfo{person}{Maarten Bosma}, \bibinfo{person}{Gaurav Mishra}, \bibinfo{person}{Adam Roberts}, \bibinfo{person}{Paul Barham}, \bibinfo{person}{Hyung~Won Chung}, \bibinfo{person}{Charles Sutton}, \bibinfo{person}{Sebastian Gehrmann}, {et~al\mbox{.}}} \bibinfo{year}{2022}\natexlab{}.
\newblock \showarticletitle{Palm: Scaling language modeling with pathways}.
\newblock \bibinfo{journal}{\emph{arXiv preprint arXiv:2204.02311}} (\bibinfo{year}{2022}).
\newblock


\bibitem[Chuang et~al\mbox{.}(2023)]%
        {chuang2023expand}
\bibfield{author}{\bibinfo{person}{Yung-Sung Chuang}, \bibinfo{person}{Wei Fang}, \bibinfo{person}{Shang-Wen Li}, \bibinfo{person}{Wen-tau Yih}, {and} \bibinfo{person}{James Glass}.} \bibinfo{year}{2023}\natexlab{}.
\newblock \showarticletitle{Expand, Rerank, and Retrieve: Query Reranking for Open-Domain Question Answering}.
\newblock \bibinfo{journal}{\emph{arXiv preprint arXiv:2305.17080}} (\bibinfo{year}{2023}).
\newblock


\bibitem[Harris(2017)]%
        {harris2017finding}
\bibfield{author}{\bibinfo{person}{Christopher~G Harris}.} \bibinfo{year}{2017}\natexlab{}.
\newblock \showarticletitle{Finding the best job applicants for a job posting: A comparison of human resources search strategies}. In \bibinfo{booktitle}{\emph{2017 IEEE International Conference on Data Mining Workshops (ICDMW)}}. IEEE, \bibinfo{pages}{189--194}.
\newblock


\bibitem[He et~al\mbox{.}(2017)]%
        {ncf}
\bibfield{author}{\bibinfo{person}{Xiangnan He}, \bibinfo{person}{Lizi Liao}, \bibinfo{person}{Hanwang Zhang}, \bibinfo{person}{Liqiang Nie}, \bibinfo{person}{Xia Hu}, {and} \bibinfo{person}{Tat-Seng Chua}.} \bibinfo{year}{2017}\natexlab{}.
\newblock \showarticletitle{Neural collaborative filtering}. In \bibinfo{booktitle}{\emph{Proceedings of the 26th international conference on world wide web}}. \bibinfo{pages}{173--182}.
\newblock


\bibitem[Hong et~al\mbox{.}(2023)]%
        {hong2023metagpt}
\bibfield{author}{\bibinfo{person}{Sirui Hong}, \bibinfo{person}{Xiawu Zheng}, \bibinfo{person}{Jonathan Chen}, \bibinfo{person}{Yuheng Cheng}, \bibinfo{person}{Jinlin Wang}, \bibinfo{person}{Ceyao Zhang}, \bibinfo{person}{Zili Wang}, \bibinfo{person}{Steven Ka~Shing Yau}, \bibinfo{person}{Zijuan Lin}, \bibinfo{person}{Liyang Zhou}, {et~al\mbox{.}}} \bibinfo{year}{2023}\natexlab{}.
\newblock \showarticletitle{Metagpt: Meta programming for multi-agent collaborative framework}.
\newblock \bibinfo{journal}{\emph{arXiv preprint arXiv:2308.00352}} (\bibinfo{year}{2023}).
\newblock


\bibitem[Kwon et~al\mbox{.}(2023)]%
        {Kwon2023EfficientMM}
\bibfield{author}{\bibinfo{person}{Woosuk Kwon}, \bibinfo{person}{Zhuohan Li}, \bibinfo{person}{Siyuan Zhuang}, \bibinfo{person}{Ying Sheng}, \bibinfo{person}{Lianmin Zheng}, \bibinfo{person}{Cody~Hao Yu}, \bibinfo{person}{Joseph~E. Gonzalez}, \bibinfo{person}{Haotong Zhang}, {and} \bibinfo{person}{Ion Stoica}.} \bibinfo{year}{2023}\natexlab{}.
\newblock \showarticletitle{Efficient Memory Management for Large Language Model Serving with PagedAttention}.
\newblock \bibinfo{journal}{\emph{Proceedings of the 29th Symposium on Operating Systems Principles}} (\bibinfo{year}{2023}).
\newblock
\urldef\tempurl%
\url{https://api.semanticscholar.org/CorpusID:261697361}
\showURL{%
\tempurl}


\bibitem[Lab(2023)]%
        {NanBeiGe}
\bibfield{author}{\bibinfo{person}{NanBeiGe~LLM Lab}.} \bibinfo{year}{2023}\natexlab{}.
\newblock \bibinfo{title}{NanBeiGe LLM}.
\newblock
\urldef\tempurl%
\url{https://github.com/Nanbeige/Nanbeige}
\showURL{%
\tempurl}


\bibitem[Le et~al\mbox{.}(2019)]%
        {ipjf}
\bibfield{author}{\bibinfo{person}{Ran Le}, \bibinfo{person}{Wenpeng Hu}, \bibinfo{person}{Yang Song}, \bibinfo{person}{Tao Zhang}, \bibinfo{person}{Dongyan Zhao}, {and} \bibinfo{person}{Rui Yan}.} \bibinfo{year}{2019}\natexlab{}.
\newblock \showarticletitle{Towards effective and interpretable person-job fitting}. In \bibinfo{booktitle}{\emph{Proceedings of the 28th ACM International Conference on Information and Knowledge Management}}. \bibinfo{pages}{1883--1892}.
\newblock


\bibitem[Li et~al\mbox{.}(2023b)]%
        {li2023camel}
\bibfield{author}{\bibinfo{person}{Guohao Li}, \bibinfo{person}{Hasan Abed Al~Kader Hammoud}, \bibinfo{person}{Hani Itani}, \bibinfo{person}{Dmitrii Khizbullin}, {and} \bibinfo{person}{Bernard Ghanem}.} \bibinfo{year}{2023}\natexlab{b}.
\newblock \showarticletitle{Camel: Communicative agents for" mind" exploration of large scale language model society}.
\newblock \bibinfo{journal}{\emph{arXiv preprint arXiv:2303.17760}} (\bibinfo{year}{2023}).
\newblock


\bibitem[Li et~al\mbox{.}(2024)]%
        {li2024does}
\bibfield{author}{\bibinfo{person}{Jia-Min Li}, \bibinfo{person}{Ruo-Xi Zhang}, \bibinfo{person}{Tung-Ju Wu}, {and} \bibinfo{person}{Mengyu Mao}.} \bibinfo{year}{2024}\natexlab{}.
\newblock \showarticletitle{How does work autonomy in human-robot collaboration affect hotel employees’ work and health outcomes? Role of job insecurity and person-job fit}.
\newblock \bibinfo{journal}{\emph{International Journal of Hospitality Management}}  \bibinfo{volume}{117} (\bibinfo{year}{2024}), \bibinfo{pages}{103654}.
\newblock


\bibitem[Li et~al\mbox{.}(2023a)]%
        {li2023ezinterviewer}
\bibfield{author}{\bibinfo{person}{Mingzhe Li}, \bibinfo{person}{Xiuying Chen}, \bibinfo{person}{Weiheng Liao}, \bibinfo{person}{Yang Song}, \bibinfo{person}{Tao Zhang}, \bibinfo{person}{Dongyan Zhao}, {and} \bibinfo{person}{Rui Yan}.} \bibinfo{year}{2023}\natexlab{a}.
\newblock \showarticletitle{EZInterviewer: To Improve Job Interview Performance with Mock Interview Generator}. In \bibinfo{booktitle}{\emph{Proceedings of the Sixteenth ACM International Conference on Web Search and Data Mining}}. \bibinfo{pages}{1102--1110}.
\newblock


\bibitem[Liu et~al\mbox{.}(2025)]%
        {liu2025mobilesteward}
\bibfield{author}{\bibinfo{person}{Yuxuan Liu}, \bibinfo{person}{Hongda Sun}, \bibinfo{person}{Wei Liu}, \bibinfo{person}{Jian Luan}, \bibinfo{person}{Bo Du}, {and} \bibinfo{person}{Rui Yan}.} \bibinfo{year}{2025}\natexlab{}.
\newblock \showarticletitle{MobileSteward: Integrating Multiple App-Oriented Agents with Self-Evolution to Automate Cross-App Instructions}. In \bibinfo{booktitle}{\emph{Proceedings of the 31st ACM SIGKDD Conference on Knowledge Discovery and Data Mining V. 1}}. \bibinfo{pages}{883--893}.
\newblock


\bibitem[Ma et~al\mbox{.}(2023)]%
        {ma2023large}
\bibfield{author}{\bibinfo{person}{Yubo Ma}, \bibinfo{person}{Yixin Cao}, \bibinfo{person}{YongChing Hong}, {and} \bibinfo{person}{Aixin Sun}.} \bibinfo{year}{2023}\natexlab{}.
\newblock \showarticletitle{Large language model is not a good few-shot information extractor, but a good reranker for hard samples!}
\newblock \bibinfo{journal}{\emph{arXiv preprint arXiv:2303.08559}} (\bibinfo{year}{2023}).
\newblock


\bibitem[Madaan et~al\mbox{.}(2023)]%
        {madaan2023self}
\bibfield{author}{\bibinfo{person}{Aman Madaan}, \bibinfo{person}{Niket Tandon}, \bibinfo{person}{Prakhar Gupta}, \bibinfo{person}{Skyler Hallinan}, \bibinfo{person}{Luyu Gao}, \bibinfo{person}{Sarah Wiegreffe}, \bibinfo{person}{Uri Alon}, \bibinfo{person}{Nouha Dziri}, \bibinfo{person}{Shrimai Prabhumoye}, \bibinfo{person}{Yiming Yang}, {et~al\mbox{.}}} \bibinfo{year}{2023}\natexlab{}.
\newblock \showarticletitle{Self-refine: Iterative refinement with self-feedback}.
\newblock \bibinfo{journal}{\emph{arXiv preprint arXiv:2303.17651}} (\bibinfo{year}{2023}).
\newblock


\bibitem[Malinowski et~al\mbox{.}(2006)]%
        {malinowski2006matching}
\bibfield{author}{\bibinfo{person}{Jochen Malinowski}, \bibinfo{person}{Tobias Keim}, \bibinfo{person}{Oliver Wendt}, {and} \bibinfo{person}{Tim Weitzel}.} \bibinfo{year}{2006}\natexlab{}.
\newblock \showarticletitle{Matching people and jobs: A bilateral recommendation approach}. In \bibinfo{booktitle}{\emph{Proceedings of the 39th Annual Hawaii International Conference on System Sciences (HICSS'06)}}, Vol.~\bibinfo{volume}{6}. IEEE, \bibinfo{pages}{137c--137c}.
\newblock


\bibitem[OpenAI(2023)]%
        {gpt4}
\bibfield{author}{\bibinfo{person}{OpenAI}.} \bibinfo{year}{2023}\natexlab{}.
\newblock \showarticletitle{GPT-4 Technical Report}.
\newblock \bibinfo{journal}{\emph{arXiv preprint arXiv:2303.08774}} (\bibinfo{year}{2023}).
\newblock


\bibitem[Ouyang et~al\mbox{.}(2022a)]%
        {instructgpt}
\bibfield{author}{\bibinfo{person}{Long Ouyang}, \bibinfo{person}{Jeffrey Wu}, \bibinfo{person}{Xu Jiang}, \bibinfo{person}{Diogo Almeida}, \bibinfo{person}{Carroll Wainwright}, \bibinfo{person}{Pamela Mishkin}, \bibinfo{person}{Chong Zhang}, \bibinfo{person}{Sandhini Agarwal}, \bibinfo{person}{Katarina Slama}, \bibinfo{person}{Alex Ray}, {et~al\mbox{.}}} \bibinfo{year}{2022}\natexlab{a}.
\newblock \showarticletitle{Training language models to follow instructions with human feedback}.
\newblock \bibinfo{journal}{\emph{Advances in Neural Information Processing Systems}}  \bibinfo{volume}{35} (\bibinfo{year}{2022}), \bibinfo{pages}{27730--27744}.
\newblock


\bibitem[Ouyang et~al\mbox{.}(2022b)]%
        {ouyang2022training}
\bibfield{author}{\bibinfo{person}{Long Ouyang}, \bibinfo{person}{Jeffrey Wu}, \bibinfo{person}{Xu Jiang}, \bibinfo{person}{Diogo Almeida}, \bibinfo{person}{Carroll Wainwright}, \bibinfo{person}{Pamela Mishkin}, \bibinfo{person}{Chong Zhang}, \bibinfo{person}{Sandhini Agarwal}, \bibinfo{person}{Katarina Slama}, \bibinfo{person}{Alex Ray}, {et~al\mbox{.}}} \bibinfo{year}{2022}\natexlab{b}.
\newblock \showarticletitle{Training language models to follow instructions with human feedback}.
\newblock \bibinfo{journal}{\emph{Advances in Neural Information Processing Systems}}  \bibinfo{volume}{35} (\bibinfo{year}{2022}), \bibinfo{pages}{27730--27744}.
\newblock


\bibitem[Qin et~al\mbox{.}(2023)]%
        {qin2023chatgpt}
\bibfield{author}{\bibinfo{person}{Chengwei Qin}, \bibinfo{person}{Aston Zhang}, \bibinfo{person}{Zhuosheng Zhang}, \bibinfo{person}{Jiaao Chen}, \bibinfo{person}{Michihiro Yasunaga}, {and} \bibinfo{person}{Diyi Yang}.} \bibinfo{year}{2023}\natexlab{}.
\newblock \showarticletitle{Is ChatGPT a general-purpose natural language processing task solver?}
\newblock \bibinfo{journal}{\emph{arXiv preprint arXiv:2302.06476}} (\bibinfo{year}{2023}).
\newblock


\bibitem[Qin et~al\mbox{.}(2018)]%
        {apjfnn}
\bibfield{author}{\bibinfo{person}{Chuan Qin}, \bibinfo{person}{Hengshu Zhu}, \bibinfo{person}{Tong Xu}, \bibinfo{person}{Chen Zhu}, \bibinfo{person}{Liang Jiang}, \bibinfo{person}{Enhong Chen}, {and} \bibinfo{person}{Hui Xiong}.} \bibinfo{year}{2018}\natexlab{}.
\newblock \showarticletitle{Enhancing person-job fit for talent recruitment: An ability-aware neural network approach}. In \bibinfo{booktitle}{\emph{The 41st international ACM SIGIR conference on research \& development in information retrieval}}. \bibinfo{pages}{25--34}.
\newblock


\bibitem[Radford et~al\mbox{.}(2019)]%
        {gpt2}
\bibfield{author}{\bibinfo{person}{Alec Radford}, \bibinfo{person}{Jeffrey Wu}, \bibinfo{person}{Rewon Child}, \bibinfo{person}{David Luan}, \bibinfo{person}{Dario Amodei}, \bibinfo{person}{Ilya Sutskever}, {et~al\mbox{.}}} \bibinfo{year}{2019}\natexlab{}.
\newblock \showarticletitle{Language models are unsupervised multitask learners}.
\newblock \bibinfo{journal}{\emph{OpenAI blog}} \bibinfo{volume}{1}, \bibinfo{number}{8} (\bibinfo{year}{2019}), \bibinfo{pages}{9}.
\newblock


\bibitem[Raffel et~al\mbox{.}(2020)]%
        {t5}
\bibfield{author}{\bibinfo{person}{Colin Raffel}, \bibinfo{person}{Noam Shazeer}, \bibinfo{person}{Adam Roberts}, \bibinfo{person}{Katherine Lee}, \bibinfo{person}{Sharan Narang}, \bibinfo{person}{Michael Matena}, \bibinfo{person}{Yanqi Zhou}, \bibinfo{person}{Wei Li}, {and} \bibinfo{person}{Peter~J Liu}.} \bibinfo{year}{2020}\natexlab{}.
\newblock \showarticletitle{Exploring the limits of transfer learning with a unified text-to-text transformer}.
\newblock \bibinfo{journal}{\emph{The Journal of Machine Learning Research}} \bibinfo{volume}{21}, \bibinfo{number}{1} (\bibinfo{year}{2020}), \bibinfo{pages}{5485--5551}.
\newblock


\bibitem[Rendle et~al\mbox{.}(2012)]%
        {rendle2012bpr}
\bibfield{author}{\bibinfo{person}{Steffen Rendle}, \bibinfo{person}{Christoph Freudenthaler}, \bibinfo{person}{Zeno Gantner}, {and} \bibinfo{person}{Lars Schmidt-Thieme}.} \bibinfo{year}{2012}\natexlab{}.
\newblock \showarticletitle{BPR: Bayesian personalized ranking from implicit feedback}.
\newblock \bibinfo{journal}{\emph{arXiv preprint arXiv:1205.2618}} (\bibinfo{year}{2012}).
\newblock


\bibitem[Shao et~al\mbox{.}(2023)]%
        {shao2023character}
\bibfield{author}{\bibinfo{person}{Yunfan Shao}, \bibinfo{person}{Linyang Li}, \bibinfo{person}{Junqi Dai}, {and} \bibinfo{person}{Xipeng Qiu}.} \bibinfo{year}{2023}\natexlab{}.
\newblock \showarticletitle{Character-llm: A trainable agent for role-playing}.
\newblock \bibinfo{journal}{\emph{arXiv preprint arXiv:2310.10158}} (\bibinfo{year}{2023}).
\newblock


\bibitem[Shen et~al\mbox{.}(2018)]%
        {jlmia}
\bibfield{author}{\bibinfo{person}{Dazhong Shen}, \bibinfo{person}{Hengshu Zhu}, \bibinfo{person}{Chen Zhu}, \bibinfo{person}{Tong Xu}, \bibinfo{person}{Chao Ma}, {and} \bibinfo{person}{Hui Xiong}.} \bibinfo{year}{2018}\natexlab{}.
\newblock \showarticletitle{A joint learning approach to intelligent job interview assessment.}. In \bibinfo{booktitle}{\emph{IJCAI}}, Vol.~\bibinfo{volume}{18}. \bibinfo{pages}{3542--3548}.
\newblock


\bibitem[Shinn et~al\mbox{.}(2023)]%
        {shinn2023reflexion}
\bibfield{author}{\bibinfo{person}{Noah Shinn}, \bibinfo{person}{Federico Cassano}, \bibinfo{person}{Beck Labash}, \bibinfo{person}{Ashwin Gopinath}, \bibinfo{person}{Karthik Narasimhan}, {and} \bibinfo{person}{Shunyu Yao}.} \bibinfo{year}{2023}\natexlab{}.
\newblock \showarticletitle{Reflexion: Language Agents with Verbal Reinforcement Learning}.
\newblock \bibinfo{journal}{\emph{arXiv preprint arXiv:2303.11366}} (\bibinfo{year}{2023}).
\newblock


\bibitem[Sun et~al\mbox{.}(2024a)]%
        {sun2024harnessing}
\bibfield{author}{\bibinfo{person}{Hongda Sun}, \bibinfo{person}{Yuxuan Liu}, \bibinfo{person}{Chengwei Wu}, \bibinfo{person}{Haiyu Yan}, \bibinfo{person}{Cheng Tai}, \bibinfo{person}{Xin Gao}, \bibinfo{person}{Shuo Shang}, {and} \bibinfo{person}{Rui Yan}.} \bibinfo{year}{2024}\natexlab{a}.
\newblock \showarticletitle{Harnessing Multi-Role Capabilities of Large Language Models for Open-Domain Question Answering}. In \bibinfo{booktitle}{\emph{Proceedings of the ACM on Web Conference 2024}}. \bibinfo{pages}{4372--4382}.
\newblock


\bibitem[Sun et~al\mbox{.}(2024b)]%
        {sun2024determlr}
\bibfield{author}{\bibinfo{person}{Hongda Sun}, \bibinfo{person}{Weikai Xu}, \bibinfo{person}{Wei Liu}, \bibinfo{person}{Jian Luan}, \bibinfo{person}{Bin Wang}, \bibinfo{person}{Shuo Shang}, \bibinfo{person}{Ji-Rong Wen}, {and} \bibinfo{person}{Rui Yan}.} \bibinfo{year}{2024}\natexlab{b}.
\newblock \showarticletitle{DetermLR: Augmenting LLM-based Logical Reasoning from Indeterminacy to Determinacy}. In \bibinfo{booktitle}{\emph{Proceedings of the 62nd Annual Meeting of the Association for Computational Linguistics (Volume 1: Long Papers)}}. \bibinfo{pages}{9828--9862}.
\newblock


\bibitem[Wang et~al\mbox{.}(2023)]%
        {wang2023rolellm}
\bibfield{author}{\bibinfo{person}{Zekun~Moore Wang}, \bibinfo{person}{Zhongyuan Peng}, \bibinfo{person}{Haoran Que}, \bibinfo{person}{Jiaheng Liu}, \bibinfo{person}{Wangchunshu Zhou}, \bibinfo{person}{Yuhan Wu}, \bibinfo{person}{Hongcheng Guo}, \bibinfo{person}{Ruitong Gan}, \bibinfo{person}{Zehao Ni}, \bibinfo{person}{Man Zhang}, {et~al\mbox{.}}} \bibinfo{year}{2023}\natexlab{}.
\newblock \showarticletitle{Rolellm: Benchmarking, eliciting, and enhancing role-playing abilities of large language models}.
\newblock \bibinfo{journal}{\emph{arXiv preprint arXiv:2310.00746}} (\bibinfo{year}{2023}).
\newblock


\bibitem[Yan et~al\mbox{.}(2019)]%
        {yan2019interview}
\bibfield{author}{\bibinfo{person}{Rui Yan}, \bibinfo{person}{Ran Le}, \bibinfo{person}{Yang Song}, \bibinfo{person}{Tao Zhang}, \bibinfo{person}{Xiangliang Zhang}, {and} \bibinfo{person}{Dongyan Zhao}.} \bibinfo{year}{2019}\natexlab{}.
\newblock \showarticletitle{Interview choice reveals your preference on the market: To improve job-resume matching through profiling memories}. In \bibinfo{booktitle}{\emph{Proceedings of the 25th ACM SIGKDD international conference on knowledge discovery \& data mining}}. \bibinfo{pages}{914--922}.
\newblock


\bibitem[Zhao et~al\mbox{.}(2022b)]%
        {zhao2022revisiting}
\bibfield{author}{\bibinfo{person}{Wayne~Xin Zhao}, \bibinfo{person}{Zihan Lin}, \bibinfo{person}{Zhichao Feng}, \bibinfo{person}{Pengfei Wang}, {and} \bibinfo{person}{Ji-Rong Wen}.} \bibinfo{year}{2022}\natexlab{b}.
\newblock \showarticletitle{A revisiting study of appropriate offline evaluation for top-N recommendation algorithms}.
\newblock \bibinfo{journal}{\emph{ACM Transactions on Information Systems}} \bibinfo{volume}{41}, \bibinfo{number}{2} (\bibinfo{year}{2022}), \bibinfo{pages}{1--41}.
\newblock


\bibitem[Zhao et~al\mbox{.}(2022a)]%
        {zhao2022tencentpretrain}
\bibfield{author}{\bibinfo{person}{Zhe Zhao}, \bibinfo{person}{Yudong Li}, \bibinfo{person}{Cheng Hou}, \bibinfo{person}{Jing Zhao}, \bibinfo{person}{Rong Tian}, \bibinfo{person}{Weijie Liu}, \bibinfo{person}{Yiren Chen}, \bibinfo{person}{Ningyuan Sun}, \bibinfo{person}{Haoyan Liu}, \bibinfo{person}{Weiquan Mao}, {et~al\mbox{.}}} \bibinfo{year}{2022}\natexlab{a}.
\newblock \showarticletitle{TencentPretrain: A Scalable and Flexible Toolkit for Pre-training Models of Different Modalities}.
\newblock \bibinfo{journal}{\emph{arXiv preprint arXiv:2212.06385}} (\bibinfo{year}{2022}).
\newblock


\bibitem[Zhu et~al\mbox{.}(2018a)]%
        {pjfnn}
\bibfield{author}{\bibinfo{person}{Chen Zhu}, \bibinfo{person}{Hengshu Zhu}, \bibinfo{person}{Hui Xiong}, \bibinfo{person}{Chao Ma}, \bibinfo{person}{Fang Xie}, \bibinfo{person}{Pengliang Ding}, {and} \bibinfo{person}{Pan Li}.} \bibinfo{year}{2018}\natexlab{a}.
\newblock \showarticletitle{Person-job fit: Adapting the right talent for the right job with joint representation learning}.
\newblock \bibinfo{journal}{\emph{ACM Transactions on Management Information Systems (TMIS)}} \bibinfo{volume}{9}, \bibinfo{number}{3} (\bibinfo{year}{2018}), \bibinfo{pages}{1--17}.
\newblock


\bibitem[Zhu et~al\mbox{.}(2018b)]%
        {zhu2018person}
\bibfield{author}{\bibinfo{person}{Chen Zhu}, \bibinfo{person}{Hengshu Zhu}, \bibinfo{person}{Hui Xiong}, \bibinfo{person}{Chao Ma}, \bibinfo{person}{Fang Xie}, \bibinfo{person}{Pengliang Ding}, {and} \bibinfo{person}{Pan Li}.} \bibinfo{year}{2018}\natexlab{b}.
\newblock \showarticletitle{Person-job fit: Adapting the right talent for the right job with joint representation learning}.
\newblock \bibinfo{journal}{\emph{ACM Transactions on Management Information Systems (TMIS)}} \bibinfo{volume}{9}, \bibinfo{number}{3} (\bibinfo{year}{2018}), \bibinfo{pages}{1--17}.
\newblock


\end{thebibliography}

\appendix

\section{Implementation Details}

In the proposed framework, we take advantage of the role-playing capabilities of LLMs, forming augmented mock interviews to enhance the evaluation process. 
We employ the open-sourced general LLM \texttt{Nanbeige-16b-chat}~\cite{NanBeiGe} as our backbone to play the roles of interviewers and candidates. We use the \texttt{vLLM}~\cite{Kwon2023EfficientMM} as the inference framework to make LLMs efficiently generate outputs.
We set the hyper-parameters in LLMs as follows: temperature=0.9, top\_p=0.6, repetition\_{penalty}=1.1, and max\_tokens=32,768.
We use the training set to update and accumulate reflection memory, and then directly perform inference on the test set without any additional fine-tuning.
For the implementation of baseline models, we uniformly use the default fine-tuning hyper-parameters for TecentPretrain~\cite{zhao2022tencentpretrain} and K2R~\cite{adolphs2021reason} and set the learning rate as 0.001.

\section{Prompt Templates}
\subsection*{Interview question raising ($f_{ques}$)}
\begin{quote}
    \texttt{Based on the interviewer you play, ask the candidate a technical question based on the Resume, Job Description, and Interview Dialogue Context (if any). The question must be closely related to the Resume/JD Content, about project experience and related skills. The questions are not allowed to be the same as before, please be as concise as possible.}
\end{quote}

\subsection*{Interview response generation ($g_{resp}$)}
\begin{quote}
    \texttt{Based on the job seeker you play, answer the interviewer's last question based on the Resume, Job Description, and Interview Dialogue Context. The answer is not allowed to be the same as before and does not conflict with the content of the applicant's resume, please be as concise as possible.}
\end{quote}

\subsection*{Interview performance evaluation ($f_{eval}$)}
\begin{quote}
    \texttt{Based on the interviewer you play, judge whether the candidate is suitable for this position based on the Resume, Job Description, and Interview Dialogue Context History. First, pay attention to whether the Job description matches the Resume and give a matching score. Then, combine the interview performance with a comprehensive matching score.}
\end{quote}

\subsection*{Job position evaluation ($g_{eval}$)}
\begin{quote}
    \texttt{Based on the job seeker you play, judge whether you can get this position based on the Resume, Job Description, and Interview Dialogue Context History. First, pay attention to whether the Job description matches your Resume and give a matching score. Then, combine your interview performance with a comprehensive matching score.}
\end{quote}

\end{document}